%% file: main.tex
\documentclass{article}

\usepackage[nonatbib,final]{neurips_2024}

\usepackage[utf8]{inputenc} 
\usepackage[T1]{fontenc}    
\usepackage{hyperref}       
\usepackage{url}            
\usepackage{booktabs}       
\usepackage{amsfonts}       
\usepackage{nicefrac}       
\usepackage{microtype}      
\usepackage{xcolor}         

\usepackage{amsmath}
\usepackage{amssymb}
\usepackage{colortbl}
\usepackage{times}
\usepackage{epsfig}
\usepackage{soul}
\usepackage{subcaption}
\usepackage{bbold}
\usepackage[mathletters]{ucs}

\definecolor{oblue}{HTML}{0A3161}
\definecolor{ored}{HTML}{B31942}

\colorlet{citecolor}{ored}

\colorlet{improvement}{oblue}
\colorlet{regression}{ored!90!black}

\usepackage[capitalize]{cleveref}
\crefname{section}{Sec.}{Secs.}
\Crefname{section}{Section}{Sections}
\Crefname{table}{Table}{Tables}
\crefname{table}{Tab.}{Tabs.}
\Crefname{appendix}{Appendix}{Appendices}

\DeclareMathAlphabet\mathbfcal{OMS}{cmsy}{b}{n}
\newcommand{\natten}{$\mathcal{N}ATTEN$}

\newcommand{\bigO}{\mathcal{O}}

\newcommand{\pc}[1]{
    \ifthenelse{#1 > 0}{
        \textcolor{improvement}{$\boldsymbol{\uparrow}$ \textbf{#1 \%}}
    }{
        \ifthenelse{#1 = 0}{
            \textbf{0 \%}
        }{
            \textcolor{regression}{$\boldsymbol{\downarrow}$ \textbf{#1 \%}}
        }
    }
}
\newcommand{\ws}[1]{#1 \texttimes{} #1}

\title{
Faster Neighborhood Attention: Reducing the $\bigO(n^2)$ Cost of Self Attention at the Threadblock Level
}

\author{
    Ali Hassani$^1$, Wen-mei Hwu$^{2,3}$, Humphrey Shi$^{1,3}$\\
    {{\small $^1$SHI Labs @ Georgia Tech, $^2$NVIDIA, $^3$UIUC}}
}

\begin{document}

\maketitle

\begin{abstract}

Neighborhood attention reduces the cost of self attention by restricting each token's attention span
to its nearest neighbors.
This restriction, parameterized by a window size and dilation factor, draws a spectrum of possible attention patterns between 
linear projection and self attention.
Neighborhood attention, and more generally sliding window attention patterns, have long been bounded by infrastructure, 
particularly in higher-rank spaces (2-D and 3-D), 
calling for the development of custom kernels, which have been limited in either functionality, or performance, if not both.
In this work, we aim to massively improve upon existing infrastructure by providing two new methods for implementing
neighborhood attention.
We first show that neighborhood attention can be represented as a batched GEMM problem, similar to standard attention, 
and implement it for 1-D and 2-D neighborhood attention.
These kernels on average provide 895\% and 272\% improvement in full precision runtime compared to existing naive CUDA kernels 
for 1-D and 2-D neighborhood attention respectively.
We find that aside from being heavily bound by memory bandwidth, certain inherent inefficiencies exist in all unfused implementations of neighborhood attention, which in most cases undo their theoretical efficiency gain.
Motivated by the progress made into fused dot-product attention kernels, we developed fused neighborhood attention;
an adaptation of fused dot-product attention kernels that allow fine-grained control over attention across different 
spatial axes.
Known for reducing the quadratic time complexity of self attention to a linear complexity, neighborhood attention can now enjoy
a reduced and constant memory footprint, and record-breaking half precision runtime.
We observe that our fused implementation successfully circumvents some of the unavoidable inefficiencies in unfused
implementations.
While our unfused GEMM-based kernels only improve half precision performance compared to naive kernels by an average of 548\% and 193\% in 
1-D and 2-D problems respectively,
our fused kernels improve naive kernels by an average of 1759\% and 958\% in 1-D and 2-D problems respectively.
These improvements translate into up to 104\% improvement in inference and 39\% improvement in training existing models based on neighborhood attention,
and additionally extend its applicability to image and video perception, as well as other modalities.
Our work is open-sourced at \url{https://github.com/SHI-Labs/NATTEN/}.
  
\end{abstract}

\begin{figure}[ht!]
    \centering
    \includegraphics[width=\textwidth]{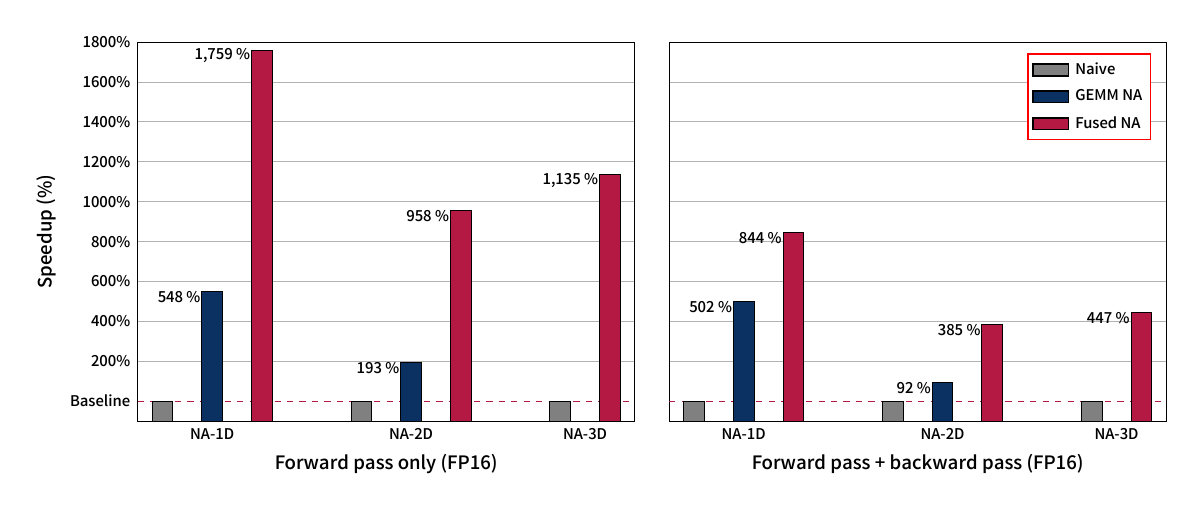}
    \caption{
        \textbf{Overview of average improvement in speed on A100 from our proposed implementation.}
        Baseline is the set of naive CUDA kernels introduced in Neighborhood Attention Transformer~\cite{hassani2023neighborhood}.
        GEMM-based NA improves
        1-D problems by an average of 548\% (forward pass) and 502\% (forward + backward), and 
        2-D problems by an average of 193\% (forward pass) and 92\% (forward + backward).
        GEMM-based NA does not implement 3-D problems yet.
        Fused NA boosts performance further and improves 
        1-D problems by an average of 1759\% (forward pass) and 844\% (forward + backward), and 
        2-D problems by an average of 958\% (forward pass) and 385\% (forward + backward), and 
        3-D problems by an average of 1135\% (forward pass) and 447\% (forward + backward).
    }
    \label{fig:improvement-plot}
\end{figure}

\section{Introduction}
Inarguably among the most highly utilized and influential primitives in modern deep learning, attention has long been cited for
its complexity and memory footprint, especially when the query and context sets are identical (self attention).
For years since its adoption in deep learning~\cite{vaswani2017attention}, the most common implementation of attention was 
through two batched GEMM (General Matrix-Matrix Multiplication) operations, sometimes referred to as ``BMM-style'' attention.
This implementation stores attention weights to global memory, which can become a bottleneck in both speed and
memory footprint. As the number of tokens grow, the number of attention weights grow as well, and the problem gets bounded by global memory bandwidth and capacity.

Over the past few years, some works proposed attention implementations in which attention weights are
kept in on-chip memory (shared memory or register file) instead,
until the second matrix multiplication is performed and the resulting attention outputs are written directly to global memory
~\cite{rabe2021self,dao2022flashattention}.
These implementations, known as fused or memory-efficient attention, reduce the number of global memory accesses in addition to global memory usage,
and successfuly turn dot product attention into a compute-bound problem at scale. 
Thanks to the first open-source implementation, Flash Attention~\cite{dao2022flashattention},
these fused attention kernels have started replacing the standard BMM-style implementations in many deep learning frameworks
and inference engines such as PyTorch~\cite{paszke2019pytorch}.

Orthogonal to these efforts, many have sought to address the quadratic complexity of self attention, which can become a significant
bottleneck in vision models more quickly. Neighborhood attention~\cite{hassani2023neighborhood} is one such method in which each query 
token is restricted to only interact with its nearest neighboring context tokens. In most cases, this pattern creates a sliding
window pattern, like that of the discrete convolution operator heavily employed in vision models.
This restriction can similarly be parameterized by a window size and dilation factor, 
and reduces the quadratic complexity of self attention down to a linear complexity.
This approach is, however, very difficult to implement at the tensor library or deep learning framework level.
Tensor views can represent sliding window attention~\cite{ramachandran2019stand}, but not the neighborhood attention pattern.
In addition, standard GEMM implementations typically do not support such tensor views in higher-rank/multi-dimensional spaces (2-D and 3-D) without explicit
copying into contiguous tensors, which in practice undoes the theoretical efficiency gain from the reduced attention complexity.
As a result, neighborhood attention was proposed along with an extension carrying naive CUDA
kernels~\cite{hassani2023neighborhood} implementing the operation.
While those kernels provided competitive FP32 performance in eager mode inference, and in some cases even FP16/BF16 performance,
they fall short of general adoption in larger scale experiments.
In addition, fused attention implementations, such as Flash Attention, effectively eliminate the $\bigO(n^2)$ memory footprint in self
attention, while also reducing runtime significantly~\cite{dao2022flashattention}, making subquadratic attention
patterns that are only possible to implement ``BMM-style'' less practical.

In this work, we present two new classes of neighborhood attention kernels: GEMM-based BMM-style kernels (GEMM NA), and 
fused kernels (Fused NA), which are aimed at providing significantly improved infrastructure for neighborhood attention.
We first show that neighborhood attention, and by extension sliding window attention, both of which are GEMV 
(General Matrix-Vector Multiplication) problems, can be expressed as GEMM problems with space-aware tiling and gather/scatter
fusion. This would allow implementing such attention patterns with performance-optimized GEMM primitives, which can also utilize specialized
hardware components such as NVIDIA's Tensor Cores.
We then extend the same logic to fused attention kernels by removing all assumptions that the token mode (``space'') is rank-1 (single-axis).
We write specializations that support higher-rank/multi-dimensional spaces, such as 2-D and 3-D. This, in theory, allows any fused attention
kernel to be modified to accommodate token spaces of any rank.
In addition, part of the logic is evaluated at compile time, resulting in less overhead.
Finally, the structural simplicity of the resulting fused neighborhood attention kernels allows for easily adding features such
as varying window sizes / dilation values across ranks/axes, causal masking, and more.

\begin{figure}
    \centering
    \includegraphics[width=\textwidth]{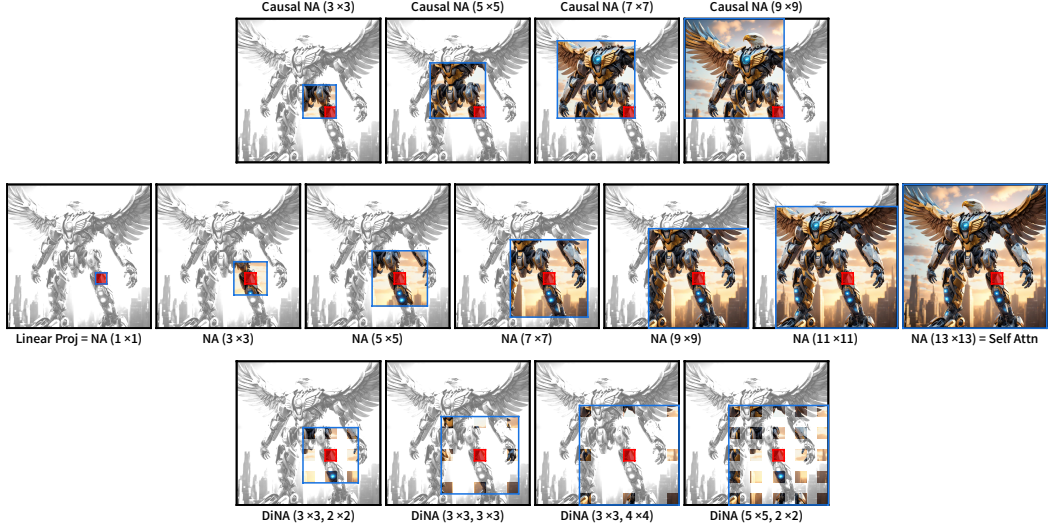}
    \caption{
    \textbf{Illustration of the spectrum of possible attention patterns provided by neighborhood attention.}
    Neighborhood attention only attempts to center the query token (red)
    within the context window (blue), unlike sliding window attention~\cite{ramachandran2019stand} which forces it.
    Neighborhood attention with window size 1 is equivalent to linear projection (``no attention'').
    Neighborhood attention approaches self attention as window size grows, and matches it when equal to input size.
    Dilation introduces sparse global context, and
    causal masking prevents interaction between query tokens that have a smaller coordinate than
    neighboring context tokens along the corresponding mode.
    Window size, dilation, and whether or not causally masked, can be defined per mode/axis.
    }
    \label{fig:neighborhoodattention}
\end{figure}

\section{Related works}
Attention being adopted as a deep learning primitive is largely owed to the Transformer architecture~\cite{vaswani2017attention}, which
despite its original application in machine translation rose to the position of being the predominant deep learning architecture.
Its design and use of the attention operator have been extended to many other applications and modalities
\cite{parmar2018image,dosovitskiy2020image,arnab2021vivit,peebles2022scalable}.
Attention is defined as an operation between a two sets of vectors: a query set and a context set.
The two undergo linear projections, with the latter projected into a set of key and value pairs. 
Scaled dot product of query and key vectors, $A$, is mapped into a probability distribution through the softmax operator,
which produces the final attention weights, $P$. The output is a set of vectors, each derived from the weighted sum of all value vectors
according to the query vector's attention weights.
It can be expressed as follows:
\begin{equation}
    Attention ( Q, K, V ) = \overbrace{ softmax \left( \underbrace{\frac{Q K^T}{\sqrt{d}}}_{A} \right) }^{P} V,
    \label{eq:attention}
\end{equation}
where $Q$, $K$, and $V$ are matrices of query, key, and value vectors as rows respectively, $\sqrt{d}$ is the scale term,
and $d$ is the number of dimensions for which the dot product is computed (number of columns in $Q$ and $K$). 
Dot product self attention, or simply, self attention, is a special case in which the query and context sets are identical.
This means for a set of $n$ input vectors, the attention weights matrix, $P$, is $\in \mathbb{R}^{n \times n}$, 
incurring an $\bigO(n^2)$ time and space complexity.
In addition, the softmax term requires column-wise reduction over the attention weight matrix, making kernel fusion more challenging.

Nevertheless, fused attention kernels successfully eliminate the $\bigO(n^2)$ global memory footprint, which makes self attention finally bound
by compute and not memory bandwidth. These two achievements paved the way for the scaling and application of attention across modalities.
To our knowledge, the first open-source implementation of a fused multi-headed attention (FMHA) kernel was contributed to the 
NVIDIA Apex \footnote{https://github.com/NVIDIA/apex} project by Young-Jun Ko, which was primarily used for accelerating
inference of Transformer-based language models. As a result of that, it was heavily limited in terms of supported models
and problem sizes, as it was performing a full softmax reduction step within the kernel.
On the other hand, Milakov and Gimelshein~\cite{milakov2018online} presented a technique for computing partial softmax statistics,
over which we can perform a final reduction step and derive exact softmax results. 
This method makes the fusion of attention kernels more practical, because they would no longer be required to compute
a full row of attention weights before proceeding to perform the second matrix multiplication.
Dao et al.~\cite{dao2022flashattention} presented and open-sourced Flash Attention, which utilizes online softmax in
order to create a performant and generic fused attention implementation.
Outperforming BMM-style implementations available in both training and inference, Flash Attention
was quickly adopted by many frameworks such as PyTorch~\cite{paszke2019pytorch}, 
and further improved for the NVIDIA Ampere~\cite{dao2023flashattention} and Hopper architectures~\cite{shah2024flashattention}.

Parallel to these efforts, many proposed restricted self attention patterns, in which context is restricted to a subset
in order to generate fewer attention weights, which in turn reduces the $\bigO(n^2)$ time and space complexity.
Stand-alone self attention (SASA)~\cite{ramachandran2019stand} is a simple 2-dimensional sliding window attention pattern, which
was shown to effectively replace convolution operations in ResNet~\cite{he2016deep} variants.
Noting challenges in implementing such patterns without incurring additional overhead from tensor copies and expansion, the
authors later moved away from explicit sliding window attention patterns to alternatives that relaxed the sliding window
movement in HaloNet~\cite{vaswani2021scaling}.
In addition to these works, sliding window attention patterns in 1-dimensional spaces has been explored in language, 
in works such as Sparse Transformers~\cite{child2019generating}, Longformer~\cite{beltagy2020longformer}, 
BigBird~\cite{zaheer2020big}, and more recently, Mistral~\cite{jiang2023mistral}. 
Neighborhood attention~\cite{hassani2023neighborhood,hassani2022dilated} is the practice of restricting the context of each
token to its nearest neighbors, which in many cases behaves like a sliding window pattern, with the exception of corner cases
in which the query cannot be centered in a sliding window. Per definitions from SASA~\cite{ramachandran2019stand}
and Longformer~\cite{beltagy2020longformer}, the sliding context window can go out of bounds, in which case the attention weights
corresponding to out-of-bounds tokens are masked. This means tokens close to spatial bounds interact with fewer context tokens.
This difference allows neighborhood attention to approach self attention as window size grows.
In addition, neighborhood attention defines a dilation factor~\cite{hassani2022dilated}, where the number of such corner cases
only increase. 
\cref{fig:neighborhoodattention} depicts possible attention patterns for a single token under different neighborhood attention
parameters.
Facing similar implementation challenges as previous works~\cite{ramachandran2019stand}, 
neighborhood attention was implemented with naive CUDA kernels packaged as a PyTorch extension, named \natten{}.
While those kernels have accelerated research in this direction, they were simply not intended to fully utilize the underlying
hardware.
The only exception is the tiled kernels, which are somewhat better optimized, but only apply to a fraction of common use cases, and
are not extensible.
In addition, with the rise of fused attention kernels such as Flash Attention~\cite{dao2022flashattention}, such implementations 
which are not performance-optimized and heavily memory-bandwidth-bound, can hardly compete in terms of 
performance and memory footprint.

To address these challenges, we present two new implementations and integrate them into \natten{}, aiming to accelerate all
neighborhood attention applications, reduce their existing memory overhead, and extend existing functionality.
We first simplify the operations that implement neighborhood attention's forward and backward pass into 3 primary operators,
and show each can be implemented as batched GEMM kernels with a fused gather/scatter operation.
We then point out key limitations in unfused neighborhood attention implementations that would prevent them from achieving 
competitive performance compared to standard BMM-style attention implementations (in more memory-bandwidth-bound cases.)
Motivated by this, and the progress made in fused attention kernels, we propose fused neighborhood attention,
which directly extends our batched GEMM methodology.
Since our main objectives are efficiency, Tensor Core utilization, and performance optimization, and
our methodology requires significant flexibility in the programming model,
we implement both approaches in CUDA C++ using NVIDIA's CUTLASS~\cite{thakkar2023cutlass} framework.
We show that the batched GEMM kernels can successfully
outperform most existing \natten{} kernels in performance, and that our fused kernels can outperform our batched GEMM kernels
while reducing the memory footprint.

\section{Methodology}

\begin{figure*}[t]
    \centering
    \includegraphics[width=\textwidth]{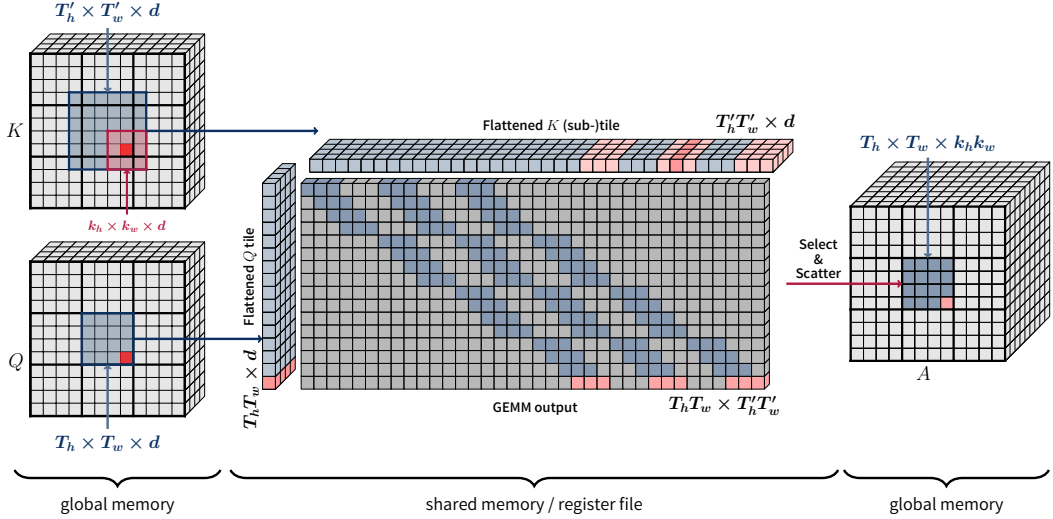}
    \caption{
    \textbf{Illustration of our GEMM-based implementation of the 2-D PN operation.} 
    Input tensors $Q$ and $K$ are tiled according to their 2-D spatial layout.
    $Q$ is tiled with a static tile shape, $T_h \times T_w$.
    $K$ is tiled with a haloed shape of the $Q$ tile, $T^{\prime}_h \times T^{\prime}_w$,
    which is a function of the attention window size ($k_h \times k_w$) and the $Q$ tile coordinates.
    Once tiles are moved into local memory, they are viewed in matrix layout,
    and a $T_h T_w \times T^{\prime}_h T^{\prime}_w \times d$ shaped GEMM is computed ($d$ is embedding dim).
    Once done, the tile of dot products with shape $T_h T_w \times T^{\prime}_h T^{\prime}_w$ is scattered into
    valid attention weights of shape $T_h \times T_w \times k_h k_w$.
    }
    \label{fig:batchedgemmna}
\end{figure*}

Herein we describe three primary operations (excluding softmax) that are required to implement a full neighborhood attention
forward and backward pass. We then show that each operation can be expressed as a batched GEMM problem, as long as tiling is
done according to the underlying spatial rank, and attention weights are scatter/gathered.
However, we find that scatter/gather is a major bottleneck for all unfused implementations of neighborhood attention, limiting
their low-precision performance specifically on more recent architectures (Ampere and later.)
We then introduce our fused neighborhood attention (FNA) formulation, which builds on our batched GEMM formulation and tiles
according to the underlying spatial rank. This approach no longer requires scatter/gathering of attention weights to/from
global memory by definition, and thereby circumvents the aforementioned bottleneck and successfully boosts
lower-precision performance on modern architectures.

\subsection{Operators}
A standard BMM-style attention forward pass (excluding softmax) is comprised of two operations: 
$QK^T$, which produces pre-softmax attention weights ($A$), and $PV$, which applies post-softmax attention weights ($P$) to
values ($V$). 
These operations are different due to layout differences in the matrix multiplications (note that $K$ is transposed, $V$ is not).
\footnote{$QK^T$ is a TN-layout and $PV$ is a TT-layout GEMM in BLAS.}

In the case of neighborhood attention, and sliding window attention in general, these will become
General Matrix-Vector Multiplication (GEMV) problems. 
In $QK^T$, each query token (vector) is multiplied by its neighboring or surrounding key tokens (matrix), and in $PV$, 
the set of attention weights corresponding to each query token (vector) is multiplied by corresponding value tokens (matrix).
Given that some of these operations can be reused in the backward pass,
we dub the $QK^T$ operation ``Pointwise-Neighborhood'' (PN) and the $PV$ operation ``Neighborhood-Neighborhood'' (NN).
PN can compute the gradient for post-softmax attention weights ($\nabla P$) when 
operating on the output gradient instead of $Q$, and $V$ instead of $K$.
Similarly, NN can compute the gradient for $Q$ ($\nabla Q$)
when operating on the pre-softmax attention gradient ($\nabla A$) instead of $A$ and $K$ instead of $V$.
We define a third operator, which can compute gradients for both $K$ and $V$: Inverse-Neighborhood (IN).
This operation is very similar to NN, but differs in gather pattern, as well as the number of attention weights. 
IN may require loading more attention weights for every token, because unlike in self attention, 
the relationship between query and context tokens in neighborhood attention is not commutative. 
In other words, query token at coordinate $i$ attending to context token at coordinate $j$ does not imply that 
query token at coordinate $j$ attends to context token at coordinate $i$.

BMM-style implementations of standard self attention have a clear edge over neighborhood and sliding window attention
implementations, because they are GEMM problems and by extension not as bound by memory bandwidth as the latter, all of which are
GEMV problems. In addition, GEMV problems cannot effectively utilize matrix multiply and accumulate (MMA) accelerators, such as
Tensor Cores.
We aim to minimize this issue by formulating all three operators as batched GEMM problems with scatter/gather fusion, in order to better utilize modern
hardware accelerators.

\subsection{Batched GEMM NA}
\label{subsec:gemm}
We transform the aforementioned GEMV problems into batched GEMMs with scatter/gather fusion.
At an abstract level, implementations of GEMM-based neighborhood attention predicate the execution of tiled MMAs
on whether any of the rows in the query tile interact with at least one of the rows in the context tile, given
the context window size, dilation, and other masking-related parameters.
We propose modifying a CUTLASS GEMM as follows in order to implement PN, NN, and IN:

\begin{enumerate}
    \item GEMM tiling is done according to the original multi-dimensional layout of the token mode in QKV.
    For example, if the attention problem is 1-D, query and context tensors are tiled along the sequence into tiles of size 64,
    for a 2-D problem, the token mode, which is comprised of height and width, are tiled by a 2-D tiler of the same size, like 8 \texttimes{} 8.

    \item Predication logic, and global pointer iterators and accessors are modified to iterate according to the original layout in
        global memory instead of assuming a standard rank-2 matrix layout.

    \item Attention weights are required to be scattered to and gathered from global memory, which in 16-bit or lower precision cannot be copied 
    asynchronously (with \verb|LDGSTS|), which breaks pipelining in those kernels on modern architectures.
    This is because the minimum transaction size for \verb|LDGSTS| is 32 bits.
\end{enumerate}

\begin{figure*}[t]
    \centering
    \includegraphics[width=\textwidth]{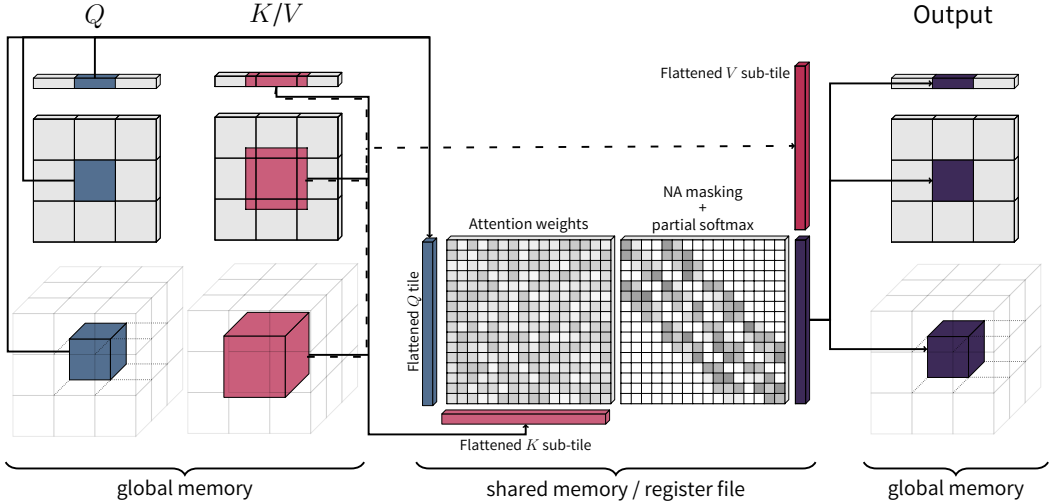}
    \caption{
        \textbf{A simplified illustration of fused neighborhood attention.} 
        $Q$ and $KV$ tensors are tiled according to their spatial layout (1-D, 2-D, 3-D), with the latter
        haloed to include the entire neighborhood for all corresponding queries in the query tile.
        Resulting attention weights from the first GEMM are masked according to neighborhood attention parameters,
        before undergoing online softmax scaling, and going through the second GEMM with the corresponding value sub-tile.
    }
    \label{fig:fna-vis}
\end{figure*}

We implemented these concepts by extending implicit GEMM (convolution) in CUTLASS (2.X API) into
kernels that compute the three neighborhood attention operators in 1-D and 2-D.
\cref{fig:batchedgemmna} shows an illustration of the 2-D GEMM-based PN kernel.
The first change is relatively inexpensive, but the second change incurs additional predication and indexing logic that can result
in additional overhead and register pressure.
The final change is a major bottleneck, and leads to lower-precision kernels (FP16/BF16) 
providing little to no improvement compared to their full precision (FP32/TF32) counterparts.
NN and IN suffer from this issue more significantly, because gathering attention weights (\verb|LDG|) breaks pipelined kernels on
Ampere, since they load GEMM operands asynchronously (\verb|LDGSTS|), which has a minimum transaction size of 32 bits.
This forces our FP16/BF16 GEMM-based kernels to fall back to global loads (\verb|LDG|), which significantly impacts achievable runtime.
To our knowledge, this issue is unavoidable in most cases, and will continue to be a bottleneck as long as attention weights are stored in
global memory.

\subsection{Fused NA}
We extend our methodology for implementing neighborhood attention operators using batched GEMM kernels to
fused attention kernels like Flash Attention~\cite{dao2022flashattention}.
This is not only motivated by the potential to reduce runtime and memory footprint, and potentially making neighborhood attention actually bound by compute,
but also to circumvent the bottleneck in the batched GEMM and naive kernels: scatter/gathering attention weights to/from global memory.
Since attention weights are only computed at the threadblock level and never fully stored in global memory in fused kernels,
the bottleneck will simply cease to exist.
We started off with xFormers FMHA~\cite{xFormers2022}, 
a fused multi-headed attention kernel based on the CUTLASS 2.X API, which can target architectures even older than Ampere
(Maxwell, SM50; Volta, SM70; and Turing, SM75.)
By carefully applying our methodology for space-aware tiling, neighborhood attention masking, and software predication for multi-dimensional tensor layouts,
we successfully implemented neighborhood attention for 1-D, 2-D, and 3-D problems.
\cref{fig:fna-vis} presents an overview of how our fused kernels function when dealing with multi-dimensional (multi-axis) data.

\subsection{Dilation and causal masking}
Our methodology allows for dilation support trivially, through simple partitioning and slicing ahead of time.
A dilated neighborhood attention problem can be mapped to a set of non-dilated neighborhood attention problems over
non-overlapping tiles of the input. All sub-problems can be computed within the same kernel call, simply by issuing more CTAs in the grid.
We additionally define and implement causal neighborhood attention into our fused kernel, which can be crucial to certain 
applications where only one spatial dimension requires causal masking (i.e. video embeddings may benefit from causally masked 
attention across the time axis and standard attention across height and width axes, which would be an exact 3-D spatio-temporal attention module.)

\subsection{Notes on arithmetic intensity}
Arithmetic intensity is the ratio of floating point operations over bytes of memory transactions, as defined by the Roofline model~\cite{williams2009roofline}:

\begin{equation}
    \text{Arithmetic Intensity} = \frac{N_{ops}}{N_{bytes}}
\end{equation}

Arithmetic intensity is typically used to determine whether an implementation/algorithm is bound by memory bandwidth or computational capacity, on a given problem size and hardware.
Let's consider a simplified representation of self attention, where we only look at pure matrix multiplication FLOPs and bytes.
Self attention is comprised of two back-to-back BMMs, which would be $2 b h n^2 d$ FLOPs for
each of the BMMs, 
where $b$, $h$, $n$, and $d$ denote batch size, number of attention heads, sequence length, and per-head dimension respectively.
In total, that would be $4 b h n^2 d$ FLOPs.
In the unfused implementation, 4 tensors with size
$b h n d$ ($Q$, $K$, $V$ and output) are accessed in global memory,
along with one intermediary tensor with size $b h n^2$ (attention weights or $P$),
which is accessed twice.
In total, that is $(4 \times b h n d + 2 \times b h n^2) \times s_{dtype}$ bytes,
where $s_{dtype}$ is the byte size of the tensor element type.
When implemented with fused attention, however, the number of bytes accessed for matrix multiplication from global memory is reduced to only reads and writes for $Q$, $K$, $V$, and attention outputs, or $4 \times b h n d \times s_{dtype}$.

Unfused implementations of attention are typically memory-bandwidth-bound at scale, given that their arithmetic intensity approaches a constant value as sequence length grows. If we take the limit of their intensity according to the aforementioned approximation of FLOPs and transaction bytes, as $n \to \infty$ with everything else as constants, we see that:
\begin{equation}
    \lim_{n \to \infty}  \frac{4 b h n^2 d}{(4 b h n d + 2 b h n^2) s_{dtype}} =
    \lim_{n \to \infty}  \frac{2 n d}{(2 d + n) s_{dtype}} =
    \frac{2 d}{s_{dtype}}
\end{equation}
Fused attention therefore solves a key problem here, by reducing the number of memory transactions from $\bigO(n^2)$ to $\bigO(n)$, which means as sequence length grows, the limit of arithmetic intensity in fused attention does not converge, and it will therefore only become bound by computational capacity.
This means that optimal fused attention kernels can almost fully utilize the underlying computational power of modern GPUs, and is the reason behind FP8 attention kernels for the Hopper architecture exceeding the 1 petaFLOP/s threshold~\cite{hippomlPetaFLOPSInference,shah2024flashattention}.
 
A natural question to ask is what happens to local attention patterns such as neighborhood attention, which promised to deliver more efficiency.
In fused implementations of neighborhood attention (i.e. our proposed FNA), 
we can look at the growth of arithmetic intensity similar to self attention. 
If we consider the FLOPs for neighborhood attention to be $4 b h n \ell d$, where $\ell$ is the size of the attention window,
and that the number of global memory transaction bytes is the same as fused self attention (worst case), $4 \times b h n d \times s_{dtype}$, then we see that as $n \to \infty$, we converge towards a constant again, therefore making neighborhood attention more memory-bandwidth-bound as $n$ alone scales:
\begin{equation}
    \lim_{n \to \infty} \frac{4 b h n \ell d}{4 b h n d s_{dtype}}
     = \frac{\ell}{s_{dtype}}
\end{equation}
However, the constant here is a function of $\ell$, the size of our attention window, which means
that as we scale the sequence length or feature map size, attention window size will determine whether or not the problem is bound by memory bandwidth or computational power.
Since smaller window sizes are closer to linear projections, and larger window sizes are closer to self attention, the fact that neighborhood attention can be bound by compute or memory bandwidth depending on window size is not a surprise.
Therefore, it is highly recommended to choose neighborhood attention window sizes according to the input size and even hardware to maximize efficiency gain.

\subsection{Limitations}
\label{subsec:na_overhead}
Our formulation of GEMM-based and fused neighborhood attention kernels poses a critical question: \textit{how much overhead can one 
expect from switching from a standard self attention kernel to neighborhood attention?}
As pointed out in \cref{subsec:gemm}, our GEMM-based kernels suffer from a major bottleneck, especially in lower-precision, 
which stems from scatter/gathering of attention weights. We consider this to be an unavoidable issue in unfused implementations of
neighborhood and sliding window attention.
Unsurprisingly, our proposed changes to fused implementations are also not free.
Changes that we find unavoidable, which in some cases can cause our fused kernels to incur higher runtime 
than the self attention baseline (xFormers FMHA) are the following (ordered by most significant to least significant):

\begin{enumerate}
    \item Kernels specialized for 2-D and 3-D problems are no longer GEMMs, they are General Tensor-Tensor contractions (GETTs)!
        Similar to convolution, if the input layout is multi-dimensional, then the GEMM
        is converted to a special case of GETT.
        On older GPU architectures, this requires more complicated software predication, which will incur more instructions and heavier register
        usage, whereas on modern architectures like Hopper, the Tensor Memory Accelerator (TMA) can easily provide hardware predication.
        Our software predication logic is similar to standard practice for such cases in CUTLASS 2.X GEMMs,
        and similarly less performant than predication for contiguous matrix layouts.
        We find this to be the most significant contributor to additional runtime in our fused kernels, when compared to the
        baseline fused self attention kernel, FMHA.
        However, FNA is perfectly capable of hiding this additional overhead in many cases, and only falls behind in cases close to
        self attention (window size is approximately the same as input size.)

    \item The attention masking logic, which depends on corresponding query and context token coordinates, original layout, and
        window size, introduces additional indexing logic in order to map linear indices to coordinates (unlike in 1-D problems
        where the mapping is the identity function), and it gets more complicated with more dimensions. 
        This, along with additional statements in the masking condition, contributes to runtime, and is expected to worsen 
        with more dimensions. Together, these contribute to more serious register spilling than the original 1-D kernel.
\end{enumerate}

Despite these issues, we find that our fused kernels can still match or outperform our self attention baseline in
approximately 100\% of 1-D, 98.6\% of 2-D, and 97.3\% of 3-D problem sizes that we benchmarked.

\section{Experiments}

\input{tables/runtime_overview.tex}

We evaluate the performance of our proposed methods by measuring their runtime against existing kernels in \natten{}.
Most use cases in \natten{} target naive CUDA kernels, with the exception of 2-D neighborhood attention with 32-dimensional
attention heads. \natten{} implements tiled kernels for those cases for up to and including window size \ws{13}, and only for
the $QK$ operation. However, we treat all kernels in \natten{} as our baseline, and will refer to them as naive kernels.
We use a fixed set of problem sizes that vary in batch size, spatial size, number of attention heads, and dimensions per 
attention head, and run them through every implementation on an NVIDIA A100 GPU and measure their runtime using CUDA events.
We iterate through multiple neighborhood attention window sizes and dilation values for every problem size.
A summary of these benchmarks is presented in \cref{tab:runtime_overview} (FP16) and \cref{tab:runtime_overview_fp32}
(FP32).
We find that our GEMM-based kernels can improve or match the naive runtime in approximately
99\% of 1-D problems (of 6150), and 
84\% of 2-D problems (of 5676) in half precision, and approximately
100\% of the 1-D problems and 
96\% of the 2-D problems in full precision.
Note that over 40\% of the 2-D problems target tiled kernels in \natten{}, which we find can sometimes
outperform our GEMM-based kernels. 
Another point of disadvantage in the FP16/BF16 variants of our GEMM-based kernels is using \verb|LDG|s in pipelined kernels, noted in \cref{subsec:gemm}.
On the other hand, our fused kernels improve or match the naive runtime in approximately
100\% of both 1-D (of 6150) and 3-D problems (of 2448) in both half precision and full precision, 
an 100\% of 2-D problems in half precision, while only improving approximately
96\% of 2-D problems in full precision.
We also find that our fused kernels match or outperform our GEMM kernels in 100\% of both 1-D and 2-D problems in half 
precision, while only doing so in approximately 65\% of 1-D problems and 74\% of 2-D problems in full precision, which is
not very surprising given that full precision is typically more memory-bandwidth-bound.
In both \cref{tab:runtime_overview} and \cref{tab:runtime_overview_fp32} we also inspect the percentage of problem sizes in
which using our fused neighborhood attention kernel is outperformed by the FMHA kernel. This is only to inspect additional
overhead caused by our implementation, which we expect to be more noticeable in 2-D and 3-D problems. Some of the overhead may
be avoidable, but our takeaway is that it is unlikely to be fully avoidable, as pointed out in \cref{subsec:na_overhead}.

We further present a breakdown of our benchmarks in \cref{tab:runtime_breakdown}, where we
report the average, minimum, and maximum improvement observed from switching from naive to GEMM-based, 
naive to fused, and GEMM-based to fused kernels.
GEMM-based kernels exhibit strong performance compared to both naive and fused kernels in full precision, where fused
kernels only have a very minimal edge over unfused.
GEMM-based kernels also outperform naive kernels in half precision, especially in cases where tiled kernels are not
available. While the tiled kernels are sometimes the better choice, we note that they simply cannot generalize to all
problem sizes as our GEMM-based kernels can, nor are they easily extensible.

\input{tables/runtime_breakdown}

\section{Future work \& Conclusion}
In this work, we formulated the neighborhood attention problem, and by extension multi-dimensional sliding window attention,
which are inherently GEMV problems, as GEMM/GETT problems.
Through this finding, we implemented extensible GEMM-based and fused CUDA kernels that implement neighborhood attention,
which can significantly improve upon existing kernels in the \natten{} project.
These kernels will not only speed up previously-proposed models based on neighborhood attention, but can also significantly
enhance ongoing research efforts in this direction.
In addition, our fused kernels are the most flexible in terms of parameterization, by supporting
varying window sizes, dilation factors, and causal masking across different axes, which enable unique
applications such as 3-D spatio-temporal attention with causal masking across time.
They also enjoy a reduced memory footprint, and can avoid being bound by memory bandwidth at scale.

Future directions in this area include but are not limited to: support for Context Parallelism (CP),
implementations using more efficient predication (i.e. with the Hopper TMA),
extension to more modern architectures (warp-specialized kernels in Hopper and Blackwell), extension to other AI accelerators,
and better auto-tuning (or alternatives involving graph compilation).

We've shown that multi-dimensional local attention can indeed serve as solutions for scaling future large-scale long-context
architectures, when provided with suitable software infrastructure.
We hope that this inspires more research into multi-dimensional attention, as deep learning systems continue to
grow larger in both model and input size.

\textbf{Acknowledgements.}
We would like to thank NVIDIA and members of the CUTLASS project, in particular Haicheng Wu, 
for his valuable feedback and comments which led to the creation of GEMM-based NA.
We also thank Meta xFormers team for developing FMHA, which is what our fused neighborhood attention kernels are based on.
A. H. thanks Michael Isaev, Aditya Kane, and Kai Wang for their feedback on the paper.
A. H. also thanks Bing Xu, Hao Lu, Michael Iovine, and Terry Chen for the invaluable learning experience while interning at HippoML,
which helped accelerate the timeline of this project.
This research was supported in part
by National Science Foundation
under Award \#2427478 - CAREER Program,
and by National Science Foundation and the Institute of Education Sciences, U.S. Department of Education
under Award \#2229873 -
National AI Institute for Exceptional Education.
This project was also partially supported by cyberinfrastructure resources and services provided by the
Partnership for an Advanced Computing Environment (PACE) at the Georgia Institute of Technology, Atlanta, Georgia, USA.

{
\small

\bibliographystyle{ieee_fullname}
\bibliography{references}
}


\clearpage

\appendix

\section{Auto-tuner}
GEMM kernels are, among other settings, parameterized by their tiling shape.
Multi-dimensional variants (2-D and 3-D neighborhood attention) can also be parameterized by
their fine-grained tile sizes, introduced by our formulation. As mentioned earlier, a GEMM with row tile size 64 can be
reinterpreted as a number of 2-D and 3-D tiles (i.e. $x \times y$ for all positive integers $x$ and $y$ where $xy = 64$, and 
$x \times y \times z$ for all positive integers $x$, $y$, and $z$ where $xyz = 64$.)
As a result, selecting tiling sizes based on factors such as problem size, hardware, and environment can
further decrease achievable runtime.
We therefore implement a very simple auto-tuning method as a proof of concept.
Auto-tuning creates and maintains a cache for the lifetime of the application,
which maps problems (defined by problem size, data type, and other such factors) to a tiling configuration.
On a cache miss, the problem is benchmarked over a set of tiling configurations, and
the best configuration gets cached.

While the auto-tuner can noticeably improve performance even further, we note that it is presently limited in the following:

\begin{enumerate}
    \item \textbf{Distributed training.} auto-tuner context is limited to a single process, meaning jobs involving distributed training or inference will run the auto-tuner separately in each individual process. Aside from the possibility of different processes choosing different settings, which can slightly impact numerical determinism, this behavior is counter-intuitive. A more advanced auto-tuner would distribute possible settings over available processes and reduce auto-tuning time in the process, and guarantee the same settings across processes.

    \item \textbf{Vast search space.} there exist in the order of thousands of valid settings for any given problem size, and searching over all of them is intractable. Our solution so far has been to generate far fewer possible settings, and even reduce the number of settings further by introducing a ``thorough mode'', which is disabled by default, but when enabled, will allow users to search over more settings and potentially gain more in speed. This issue is a common problem in modern computational packages, and we hope to alleviate it by common practices such as reducing benchmark time, distributing the process, caching to disk, lazy benchmarking, and approximate performance models.
\end{enumerate}

\section{Additional experiments}

Herein we present some additional performance metrics from our GEMM-based and fused kernels compared against the baseline.

In \cref{tab:runtime_breakdown_backward}, we break down expected performance improvements at the operation level from a single forward and backward pass. Both our GEMM-based and fused kernels provide significant improvement on average over the baseline, while there still exist cases where naive could potentially perform better, especially compared to our GEMM-based kernels. As pointed out in \cref{subsec:gemm}, the scatter and gather operation in our GEMM kernels are a significant bottleneck, especially in lower-precision and in NN and IN operations. In lower precision, NN and IN, which account for 75\% of the backward pass operations (excluding softmax) will fail to hide their prefetch runtime from global reads, which are not asynchronous, and this will essentially impact the backward pass more than it does the forward pass.
This issue, however, is limited to our unfused variant, and our fused kernels maintain their superior performance levels, offering up to an order of magnitude improvement in all variants (1-D, 2-D, and 3-D).

\input{tables/runtime_breakdown_train}

In addition to our operation-level benchmarks, we also evaluate the effect of our proposed methodology on existing models that use
neighborhood attention as a primitive, NAT~\cite{hassani2023neighborhood} and DiNAT~\cite{hassani2022dilated}.
We benchmark the throughput from all variants according to ImageNet-1K~\cite{krizhevsky2012imagenet} specifications,
and report FP16 and FP32 measurements in \cref{tab:nat_and_dinat_on_imagenet} and \cref{tab:nat_and_dinat_on_imagenet_fp32} respectively.
We also benchmark a style-based generative adversarial (GAN) model based on neighborhood attention, 
StyleNAT~\cite{walton2022stylenat} and report performance improvements in \cref{tab:stylenat_ffhq}.
We find that at least in problem sizes that the ImageNet classification models NAT and DiNAT typically require, which are
typically smaller in spatial size and window size, and larger in batch size, our GEMM-based approach fails to improve the
baseline in half precision, and only minimally improves it in full precision.
Our fused kernels on the other hand never fail to improve upon the baseline, but they only provide significant improvement
in half precision, and cases that use dilation frequently (DiNAT~\cite{hassani2022dilated} variants).
Improvements in the generative model, StyleGAN~\cite{walton2022stylenat}, are only observed in full precision (half precision
is not recommended in this application), where again we find that both our GEMM-based and fused kernels can improve inference
speed compared to existing naive kernels, with our fused kernels having a much more noticeable edge.

\input{tables/imagenet}
\input{tables/ffhq}

Finally, we also attempted to estimate improvements in training time compared to our baseline. As suggested by our earlier findings regarding the limit of our GEMM-based implementation in the backward pass, we do not see any improvement in training time compared to the naive baseline.
However, we find that our fused kernels deliver on the promise of improved half precision training time.
We present our estimates in \cref{tab:imagenet_training}, which are based on measurements from training NAT~\cite{hassani2023neighborhood} and DiNAT~\cite{hassani2022dilated} variants according to their original specifications. We ran each model for 1 warmup epoch, and 1 benchmark epoch, the average throughput of which is used to estimate training time for 300 epochs.

\input{tables/imagenet_training}


\end{document}

%% file: tables/runtime_overview.tex
\newcommand{\runtimeOverviewMaxWidth}{1.0}

\setlength{\tabcolsep}{6pt}
\begin{table}
    \begin{minipage}{0.51\columnwidth}
    \caption{
        \textbf{FP16 forward pass benchmark overview.}
        We benchmark naive neighborhood attention kernels against our proposed GEMM and fused kernels in half precision, over a
        large set of problem sizes varying in batch size, spatial size, number of attention heads, and dimensions per head, and
        over different window sizes and dilation values.
        For every problem size, we also benchmarked self attention running with the xFormers FMHA (our baseline) and 
        Flash Attention V2.
    }
    \label{tab:runtime_overview}
    \centering
    \resizebox{\runtimeOverviewMaxWidth\textwidth}{!}{
    \begin{tabular}{l|ccc|cc}
        \toprule
        \textbf{NA Kernel} & \multicolumn{5}{c}{\textbf{\% of problems matched or outperformed}} \\
        \midrule
                           & \multicolumn{3}{c|}{\textbf{neighborhood attn}}  & \multicolumn{2}{c}{\textbf{self attn}} \\
                           & \textbf{Naive} & \textbf{GEMM} & \textbf{Fused} & \textbf{FMHA} & \textbf{FAv2}  \\
        \midrule
        \multicolumn{6}{c}{\textit{1-dimensional neighborhood attention}}                                              \\
        \midrule
        \textbf{Naive}     & -              & 1.7 \%        & 0.0 \%           &  21.8 \%         &  8.8 \%          \\
        \textbf{GEMM}      &  98.7 \%       & -             & 0.0 \%           &  72.0 \%         & 54.2 \%          \\
        \textbf{Fused}     & 100.0 \%       & 100.0 \%      & -                & 100.0 \%         & 98.2 \%          \\
        \midrule
        \multicolumn{6}{c}{\textit{2-dimensional neighborhood attention}} \\
        \midrule
        \textbf{Naive}     & -                &  16.4 \%        & 0.0 \%           & 32.9 \%         & 15.8 \%          \\
        \textbf{GEMM}      & 84.0 \%          & -               & 0.0 \%           & 59.3 \%         & 29.8 \%          \\
        \textbf{Fused}     & 100.0 \%         & 100.0 \%        & -                & 98.6 \%         & 92.4 \%          \\
        \midrule
        \multicolumn{6}{c}{\textit{3-dimensional neighborhood attention}} \\
        \midrule
        \textbf{Naive}     & -               & -             & 0.0 \%         & 43.5 \%         & 20.2 \%          \\
        \textbf{Fused}     & 100.0 \%         & -             & -             & 97.3 \%         & 87.0 \%          \\
        \bottomrule
    \end{tabular}
    }

\end{minipage}
\hspace{0.015\columnwidth}
\begin{minipage}{0.47\columnwidth}

    \centering
    \caption{
        \textbf{FP32 forward pass benchmark overview.}
        We benchmark naive neighborhood attention kernels against our proposed GEMM and 
        fused kernels in full precision, over a large set of problem sizes varying in batch size, spatial size, 
        number of attention heads, and dimensions per head, and over different window sizes and dilation values.
        For every problem size, we also benchmarked self attention running with the xFormers FMHA (our baseline).
    }
    \label{tab:runtime_overview_fp32}
    \resizebox{\runtimeOverviewMaxWidth\textwidth}{!}{
    \begin{tabular}{l|ccc|c}
        \toprule
        \textbf{NA Kernel} & \multicolumn{4}{c}{\textbf{\% of problems matched or outperformed}}   \\
        \midrule
                           & \multicolumn{3}{c|}{\textbf{neighborhood attn}}  & \textbf{self attn} \\
                           & \textbf{Naive} & \textbf{GEMM} & \textbf{Fused} & \textbf{FMHA}       \\
        \midrule
        \multicolumn{5}{c}{\textit{1-dimensional neighborhood attention}}                          \\
        \midrule
        \textbf{Naive}     & -                &  0.0 \%         &  0.0 \%          & 34.6 \%               \\
        \textbf{GEMM}      &  99.9 \%         & -               & 37.7 \%          & 98.4 \%               \\
        \textbf{Fused}     & 100.0 \%         & 64.8 \%         & -                & 99.9 \%               \\
        \midrule
        \multicolumn{5}{c}{\textit{2-dimensional neighborhood attention}} \\
        \midrule
        \textbf{Naive}     & -                & 11.7 \%         &  5.4 \%          & 52.0 \%               \\
        \textbf{GEMM}      & 89.5 \%          & -               & 28.1 \%          & 92.4 \%               \\
        \textbf{Fused}     & 96.0 \%          & 74.0 \%         & -                & 99.3 \%               \\
        \midrule
        \multicolumn{5}{c}{\textit{3-dimensional neighborhood attention}} \\
        \midrule
        \textbf{Naive}     & -               & -             & 0.0 \%           & 61.1 \%               \\
        \textbf{Fused}     & 100.0 \%        & -             & -                & 98.6 \%               \\
        \bottomrule
    \end{tabular}
    }
\end{minipage}
\end{table}

%% file: tables/runtime_breakdown.tex
\newcommand{\runtimeBreakdownMaxWidth}{1.0}

\setlength{\tabcolsep}{2pt}
\begin{table}[ht!]
    \caption{
        \textbf{Forward pass benchmark breakdown.}
        Both GEMM-based and fused NA improve the baseline naive kernels on average. 
        However, there exist cases in which naive kernels may be preferable to GEMM-based in both FP16 and
        FP32, but naive is rarely a good choice in half precision where both naive and GEMM are more memory
        bandwidth bound than fused.
    }
    \label{tab:runtime_breakdown}
    \centering
    \resizebox{\runtimeBreakdownMaxWidth\textwidth}{!}{
    \begin{tabular}{l|ccc|ccc|ccc}
\toprule
\textbf{Dim}& \multicolumn{3}{c|}{\textbf{GEMM over naive}}   & \multicolumn{3}{c|}{\textbf{Fused over naive}}  & \multicolumn{3}{c}{\textbf{Fused over GEMM}} \\
          & \textbf{Average} & \textbf{Min} & \textbf{Max} & \textbf{Average} & \textbf{Min} & \textbf{Max} & \textbf{Average} & \textbf{Min} & \textbf{Max} \\

\midrule
\multicolumn{10}{c}{\textit{FP16}}\\
\midrule
\textbf{1-D}     & \pc{ 548}   &    \pc{-53}   &   \pc{3025}   &   \pc{1759}   &   \pc{  60}   &  \pc{11885}   &    \pc{180}   &   \pc{ 71}   &    \pc{466}  \\

\textbf{2-D}     &  \pc{193}   &    \pc{-57}   &    \pc{862}   &   \pc{ 958}   &   \pc{   0}   &   \pc{7169}   &    \pc{257}   &   \pc{ 38}   &   \pc{1199}  \\

\textbf{3-D}     & -           & -             & -             &   \pc{1135}   &   \pc{ 118}   &   \pc{5497}   & -             & -            & -            \\

\midrule
\multicolumn{10}{c}{\textit{FP32}}\\
\midrule

\textbf{1-D}     &  \pc{ 874}  &  \pc{ -31}  &  \pc{3565}  &  \pc{ 978}  &  \pc{  13}  &  \pc{4419}  &  \pc{ 17}  &  \pc{-54}  &  \pc{136}\\

\textbf{2-D}     &  \pc{ 386}  &  \pc{ -43}  &  \pc{1933}  &  \pc{ 564}  &  \pc{ -30}  &  \pc{4043}  &  \pc{ 43}  &  \pc{-53}  &  \pc{451}\\

\textbf{3-D}     & -           & -           & -           &  \pc{ 712}  &  \pc{  25}  &  \pc{3029}  & - & - & - \\

\bottomrule
    \end{tabular}
    }
\end{table}

%% file: tables/runtime_breakdown_train.tex
\setlength{\tabcolsep}{2pt}
\begin{table}[ht!]
    \caption{
        \textbf{Forward + backward pass benchmark breakdown.}
        Improvements over naive, while not as significant as in the forward pass, are still significant.
        We report benchmark the full forward and backward pass in half precision only, because most
        training is done in lower precision.
    }
    \label{tab:runtime_breakdown_backward}
    \centering
    \resizebox{\runtimeBreakdownMaxWidth\textwidth}{!}{
    \begin{tabular}{l|ccc|ccc|ccc}
\toprule
\textbf{Dim}& \multicolumn{3}{c|}{\textbf{GEMM over naive}}   & \multicolumn{3}{c|}{\textbf{Fused over naive}}  & \multicolumn{3}{c}{\textbf{Fused over GEMM}} \\
          & \textbf{Average} & \textbf{Min} & \textbf{Max} & \textbf{Average} & \textbf{Min} & \textbf{Max} & \textbf{Average} & \textbf{Min} & \textbf{Max} \\

\midrule
\textbf{1-D}     & \pc{ 502}   &    \pc{-48}   &   \pc{3017}   &   \pc{ 844}   &   \pc{ -20}   &   \pc{7605}   &     \pc{57}   &   \pc{-50}   &    \pc{229}  \\

\textbf{2-D}     &  \pc{ 92}   &    \pc{-70}   &    \pc{474}   &   \pc{ 385}   &   \pc{ -61}   &   \pc{3723}   &    \pc{150}   &   \pc{-49}   &    \pc{855}  \\

\textbf{3-D}     & -           & -             & -             &   \pc{ 447}   &   \pc{ -45}   &   \pc{2824}   & -             & -            & -            \\

\bottomrule
    \end{tabular}
    }
\end{table}

%% file: tables/imagenet.tex
\setlength{\tabcolsep}{2pt}
\begin{table}[ht!]
    \caption{
        \textbf{Model-level throughput changes when using our proposed GEMM-based and fused kernels in ImageNet classification.}
        Hierarchical vision transformers NAT and DiNAT can see between 26\% to 104\% improvement in FP16 throughput on an A100
        (batch size 128) with our proposed fused kernel. Suffering from the memory alignment issue, our half precision GEMM
        kernels usually result in a much smaller improvement over naive kernels, particularly the tiled variants.
        The same measurements with FP32 precision are presented in \cref{tab:nat_and_dinat_on_imagenet_fp32}.
    }
    \label{tab:nat_and_dinat_on_imagenet}
    \centering
    \begin{tabular}{l|cc|ccc|c}
        \toprule
        \textbf{Model} & \textbf{\# of}     & \textbf{FLOPs} & \multicolumn{3}{c|}{\textbf{Throughput}}         & \textbf{Top-1}\\
                       & \textbf{Params}    &                & \textbf{Naive} & \textbf{GEMM} & \textbf{Fused} & \textbf{Accuracy}  \\
                       &  (M)               &   (G)          & \multicolumn{3}{c|}{(imgs/sec)}                  & (\%) \\
        \midrule
        \textbf{NAT-M}                               &  20 &   2.7 & 2975 & 2660 (\pc{-11}) & 3742 (\pc{ 26}) &  81.8 \\
        \textbf{DiNAT-M}                             &  20 &   2.7 & 2672 & 2548 (\pc{ -5}) & 3930 (\pc{ 47}) &  81.8 \\
        \midrule
        \textbf{DiNAT$_s$-T}                         &  28 &   4.5 & 2850 & 2504 (\pc{-12}) & 3847 (\pc{ 35}) &  81.8 \\
        \textbf{NAT-T}                               &  28 &   4.3 & 2167 & 1939 (\pc{-11}) & 2772 (\pc{ 28}) &  83.2 \\
        \textbf{DiNAT-T}                             &  28 &   4.3 & 1910 & 1845 (\pc{ -3}) & 2909 (\pc{ 52}) &  82.7 \\
        \midrule
        \textbf{DiNAT$_s$-S}                         &  50 &   8.7 & 1800 & 1571 (\pc{-13}) & 2445 (\pc{ 36}) &  83.5 \\
        \textbf{NAT-S}                               &  51 &   7.8 & 1457 & 1309 (\pc{-10}) & 1879 (\pc{ 29}) &  83.7 \\
        \textbf{DiNAT-S}                             &  51 &   7.8 & 1360 & 1313 (\pc{ -3}) & 2145 (\pc{ 58}) &  83.8 \\
        \midrule
        \textbf{DiNAT$_s$-B}                         &  88 &  15.4 & 1351 & 1178 (\pc{-13}) & 1837 (\pc{ 36}) &  83.8 \\
        \textbf{NAT-B}                               &  90 &  13.7 & 1110 &  997 (\pc{-10}) & 1448 (\pc{ 30}) &  84.3 \\
        \textbf{DiNAT-B}                             &  90 &  13.7 &  982 &  950 (\pc{ -3}) & 1517 (\pc{ 54}) &  84.4 \\
        \midrule
        \textbf{DiNAT$_s$-L}                             & 197 &   34.5 &  846 &  744 (\pc{-12}) & 1119 (\pc{ 32}) &  86.5 \\
        \textbf{DiNAT-L}                                 & 200 &   30.6 &  669 &  647 (\pc{ -3}) & 1042 (\pc{ 56}) &  86.6 \\
        \textbf{DiNAT$_s$-L\textsuperscript{(\ws{384})}} & 197 &  101.5 &  295 &  239 (\pc{-19}) &  391 (\pc{ 33}) &  87.4 \\
        \textbf{DiNAT-L\textsuperscript{(\ws{384})}}     & 200 &   92.4 &  153 &  134 (\pc{-12}) &  312 (\pc{104}) &  87.5 \\
        \bottomrule
    \end{tabular}
\end{table}

\setlength{\tabcolsep}{2pt}
\begin{table}[ht!]
    \caption{
        \textbf{Model-level throughput changes when using our proposed GEMM-based and fused kernels in ImageNet classification (full precision).}
        While fused attention kernels are not expected to have as large of an edge over BMM-style attention kernels in FP32,
        our fused kernels still happen to outperform naive kernels in full precision.
        It is also visible that our GEMM kernels can outperform naive kernels when we eliminate the memory alignment issue.
        That said, our FP32 GEMM kernels still impose a maximum alignment of 1 element on the attention weights tensor, which
        limits its ability to compete with other BMM-style attention kernels.
    }
    \label{tab:nat_and_dinat_on_imagenet_fp32}
    \centering
    \begin{tabular}{l|cc|ccc|c}
        \toprule
        \textbf{Model} & \textbf{\# of}     & \textbf{FLOPs} & \multicolumn{3}{c|}{\textbf{Throughput}}         & \textbf{Top-1}\\
                       & \textbf{Params}    &                & \textbf{Naive} & \textbf{GEMM} & \textbf{Fused} & \textbf{Accuracy}  \\
                       & (M)                & (G)            & \multicolumn{3}{c|}{(imgs/sec)}                  & (\%) \\
        \midrule
        \textbf{NAT-M}                               &  20 &   2.7 & 2416 & 2481 (\pc{ 3}) & 2658 (\pc{10}) &  81.8 \\
        \textbf{DiNAT-M}                             &  20 &   2.7 & 2217 & 2364 (\pc{ 7}) & 2905 (\pc{31}) &  81.8 \\
        \midrule
        \textbf{DiNAT$_s$-T}                         &  28 &   4.5 & 2270 & 2255 (\pc{-1}) & 2771 (\pc{22}) &  81.8 \\
        \textbf{NAT-T}                               &  28 &   4.3 & 1739 & 1802 (\pc{ 4}) & 1942 (\pc{12}) &  83.2 \\
        \textbf{DiNAT-T}                             &  28 &   4.3 & 1591 & 1706 (\pc{ 7}) & 2123 (\pc{33}) &  82.7 \\
        \midrule
        \textbf{DiNAT$_s$-S}                         &  50 &   8.7 & 1403 & 1393 (\pc{-1}) & 1717 (\pc{22}) &  83.5 \\
        \textbf{NAT-S}                               &  51 &   7.8 & 1160 & 1199 (\pc{ 3}) & 1293 (\pc{11}) &  83.7 \\
        \textbf{DiNAT-S}                             &  51 &   7.8 & 1102 & 1183 (\pc{ 7}) & 1490 (\pc{35}) &  83.8 \\
        \midrule
        \textbf{DiNAT$_s$-B}                         &  88 &  15.4 & 1020 & 1009 (\pc{-1}) & 1240 (\pc{22}) &  83.8 \\
        \textbf{NAT-B}                               &  90 &  13.7 &  867 &  897 (\pc{ 3}) &  966 (\pc{11}) &  84.3 \\
        \textbf{DiNAT-B}                             &  90 &  13.7 &  795 &  851 (\pc{ 7}) & 1059 (\pc{33}) &  84.4 \\
        \midrule
        \textbf{DiNAT$_s$-L}                             & 197 &   34.5 &  609 &  601 (\pc{-1}) &  721 (\pc{18}) &  86.5 \\
        \textbf{DiNAT-L}                                 & 200 &   30.6 &  506 &  540 (\pc{ 7}) &  669 (\pc{32}) &  86.6 \\
        \textbf{DiNAT$_s$-L\textsuperscript{(\ws{384})}} & 197 &  101.5 &  211 &  193 (\pc{-9}) &  245 (\pc{16}) &  87.4 \\
        \textbf{DiNAT-L\textsuperscript{(\ws{384})}}     & 200 &   92.4 &  116 &  115 (\pc{-1}) &  179 (\pc{54}) &  87.5 \\
        \bottomrule
    \end{tabular}
\end{table}

%% file: tables/ffhq.tex
\setlength{\tabcolsep}{4pt}
\begin{table}
    \caption{
        \textbf{Model-level throughput changes when using our proposed GEMM-based and fused kernels in style-based image generation.}
        We benchmark StyleNAT~\cite{walton2022stylenat}, a style-based generative adversarial model based on neighborhood attention under different kernels.
        We experimented with different batch sizes in order to achieve peak performance, and settled for 64 for the \ws{256} variant, and 8 for the \ws{1024}.
        StyleNAT does not recommend lower-precision, therefore these measurements are only done in FP32.
    }
    \label{tab:stylenat_ffhq}
    \centering
    \begin{tabular}{lc|ccc|c}
        \toprule
        \textbf{Dataset} & \textbf{\# of}   & \multicolumn{3}{c|}{\textbf{Throughput}}         & \textbf{FID}\\
                         & \textbf{Params}    & \textbf{Naive} & \textbf{GEMM} & \textbf{Fused} &  \\
                         &                    & \multicolumn{3}{c|}{(imgs/sec)}                  &  \\
        \midrule
        \textbf{FFHQ (\ws{256})}         & 48.9 M & 36.7 &  40.6 (\pc{11}) &  45.5 (\pc{24}) &  2.05 \\
        \textbf{FFHQ (\ws{1024})}        & 49.4 M &  8.2 &   8.5 (\pc{ 3}) &  11.5 (\pc{40}) &  4.17 \\
        \bottomrule
    \end{tabular}
\end{table}

%% file: tables/imagenet_training.tex
\setlength{\tabcolsep}{6pt}
\begin{table}[ht!]
    \caption{
        \textbf{Training time improvement when using fused neighborhood attention kernels.}
        We ran each of the classification models based on neighborhood attention for one warmup epoch and one benchmark epoch,
        all with half precision (the typical training scenario), and report the estimated training time.
        Note that these numbers exclude positional biases, as our fused backward kernel does not support it.
    }
    \label{tab:imagenet_training}
    \centering
    \begin{tabular}{l|cc|ccc}
        \toprule
        \textbf{Model} & \textbf{\# of}     & \textbf{FLOPs} & \multicolumn{3}{c}{\textbf{Training time estimate}}     \\
                       & \textbf{Params}    &                & \textbf{Naive} & \textbf{GEMM} & \textbf{Fused}   \\
                       &  (M)               &   (G)          & \multicolumn{3}{c}{(hours)}                  \\
        \midrule
        \textbf{NAT-M}                               &  20 &   2.7 & 19.4 & 20.4 (\pc{-5}) & 16.6 (\pc{17})  \\
        \textbf{DiNAT-M}                             &  20 &   2.7 & 20.4 & 21.2 (\pc{-4}) & 17.4 (\pc{17})  \\
        \midrule
        \textbf{DiNAT$_s$-T}                         &  28 &   4.5 & 21.1 & 22.0 (\pc{-4}) & 17.4 (\pc{21})  \\
        \textbf{NAT-T}                               &  28 &   4.3 & 26.5 & 28.2 (\pc{-6}) & 24.0 (\pc{10})  \\
        \textbf{DiNAT-T}                             &  28 &   4.3 & 27.4 & 28.5 (\pc{-4}) & 21.9 (\pc{25})  \\
        \midrule
        \textbf{DiNAT$_s$-S}                         &  50 &   8.7 & 33.3 & 33.2 (\pc{ 0}) & 25.1 (\pc{33})  \\
        \textbf{NAT-S}                               &  51 &   7.8 & 39.2 & 41.8 (\pc{-6}) & 33.7 (\pc{16})  \\
        \textbf{DiNAT-S}                             &  51 &   7.8 & 38.0 & 40.1 (\pc{-5}) & 30.8 (\pc{23})  \\
        \midrule
        \textbf{DiNAT$_s$-B}                         &  88 &  15.4 & 45.4 & 46.1 (\pc{-2}) & 32.6 (\pc{39})  \\
        \textbf{NAT-B}                               &  90 &  13.7 & 51.1 & 54.6 (\pc{-6}) & 47.7 (\pc{ 7})   \\
        \textbf{DiNAT-B}                             &  90 &  13.7 & 54.4 & 56.0 (\pc{-3}) & 41.0 (\pc{33})   \\
        \bottomrule
    \end{tabular}
\end{table}